\pdfoutput=1

\documentclass[11pt]{article}

\usepackage[final]{acl}

\usepackage{times}
\usepackage{latexsym}

\usepackage[T1]{fontenc}

\usepackage[utf8]{inputenc}

\usepackage{microtype}

\usepackage{inconsolata}
\usepackage{url}
\usepackage{graphicx}
\usepackage{booktabs}
\usepackage{amsmath}
\usepackage{dblfloatfix}
\usepackage[dvipsnames,table]{xcolor}
\usepackage{colortbl}
\definecolor{bestcol}{rgb}{0.78,0.89,1.0}      
\definecolor{secondcol}{rgb}{1.0,0.91,0.72}    
\usepackage{tcolorbox}
\tcbuselibrary{skins,breakable}
\usepackage{tikz}
\usepackage{multirow}
\usepackage{multicol}
\usepackage{xspace}
\usepackage{enumitem}
\setlist{topsep=2pt,itemsep=1pt,parsep=0pt}
\definecolor{c0}{cmyk}{1,0.3968,0,0.2588} 

\newtcbox{\pattern}{on line,colback=c0!10,colframe=white,size=fbox,arc=3pt, box align=base,before upper=\strut,
top=-2pt, bottom=-2pt, boxrule=0pt}
\title{SkillChain: Closing the Loop on Skill Evolution for Image-Based E-Commerce AI Assistants}

\author{
\begin{minipage}[t]{\textwidth}
\centering
\textbf{Yimin Hu\textsuperscript{1}} \quad
\textbf{Mengtao Xu\textsuperscript{1}} \quad
\textbf{Hao Guo\textsuperscript{1}} \quad
\textbf{Yuheng Song\textsuperscript{1}} \quad
\textbf{Xiaoyong Zhu\textsuperscript{1,†}} \quad
\textbf{Bo Zheng\textsuperscript{1,†}} \\
\textsuperscript{1}Taobao \& Tmall Group of Alibaba \quad
\end{minipage}
\\[2ex]
\begin{minipage}[t]{\textwidth}
\centering
\texttt{
\{hym408321, mengtao.xmt, gh225907, songyuheng.syh\}@taobao.com, \{xiaoyong.z, bozheng\}@alibaba-inc.com
}
\end{minipage}
}

\date{}

\begin{document}
\maketitle
\renewcommand{\thefootnote}{\fnsymbol{footnote}}
\footnotetext[2]{Corresponding author}
\begin{abstract}
Image-based AI assistants are now deployed at production scale on e-commerce platforms, 
where a single uploaded image can trigger fundamentally different user intents: product search,
style recommendation, visual encyclopedia, or utility tool calls, each demanding its own response 
format, tool invocation, and domain knowledge. Without per-intent behavioral constraints, 
LLM-based systems conflate these heterogeneous modes and fall short of domain quality standards, 
while the breadth and dynamism of the intent space render manual engineering infeasible.
To address this, we present \textbf{SkillChain}, which closes the production feedback
loop on Skill evolution, automating the lifecycle of \emph{Skills} through three stages:
\emph{Skill Creator} for bootstrapping from task specs and trajectories,
\emph{Route Optimizer} for routing alignment, and \emph{Body Refiner} for 
iterative Skill Body refinement via dual-path LLM-Judge evaluation.
Deployed on a production-scale e-commerce image assistant, SkillChain substantially
improves aggregate response quality, with the strongest gains on structural compliance
and content quality; a one-week online A/B experiment further confirms significant
gains in user engagement, content consumption, and long-term retention.
\end{abstract}

\section{Introduction}

E-commerce platforms increasingly deploy image-based AI assistants powered by
large language models~\citep{brown2020gpt3,openai2023gpt4,touvron2023llama2}
that allow users to upload a photo and receive personalized responses.
Visual inputs carry inherently ambiguous intent: the same image may prompt a
product search, a style comparison, an encyclopedia lookup, or a utility tool
call, each demanding a distinct response format, tool set, and domain
vocabulary.

This diversity poses three interconnected production challenges.
\textbf{(C1) No per-intent behavioral specifications.}
Without explicit constraints per intent, LLMs generate free-form responses that
mix incompatible formats such as product cards in encyclopedia replies, invoke
tools incorrectly, and fail domain quality standards.
\textbf{(C2) Routing drift under distribution shift.}
Visual intent patterns continuously evolve; well-designed intent-to-specification
mappings degrade over time, yet continuous re-engineering by hand is infeasible
at scale.
\textbf{(C3) Specification decay without production feedback.}
Specifications fixed at creation time silently accumulate deficiencies with no
mechanism for automated diagnosis and repair.

We propose \textbf{SkillChain}, which addresses C1--C3 via \emph{Skills},
declarative per-intent specifications covering tool calls, rich-media
composition, writing constraints, and domain knowledge, across three coupled
stages: \textbf{Stage~1 (Skill Creator)} bootstraps Skills from task
specifications and user trajectories, gating quality through human reflection
(C1); \textbf{Stage~2 (Route Optimizer)} mines routing failures and applies
update/merge/discard operations to realign Descriptions with evolving traffic
(C2); \textbf{Stage~3 (Body Refiner)} runs dual-path evaluation and
cross-sample attribution to identify and repair Body deficiencies (C3).

Deployed on a production-scale e-commerce image assistant, SkillChain achieves
the highest aggregate LLM Judge score across all evaluated configurations,
with substantial gains in structural compliance and content quality;
online A/B results confirm significant engagement and retention improvements
over the production Stage~2 baseline.
Critically, the pipeline is \emph{unidirectional}: each stage targets a disjoint component, so corrections never propagate backward, a property no prior skill system achieves.

Our main contributions are:
\begin{enumerate}
  \item We identify routing and behavioral drift as stage-specific
  production Skill failure modes, each addressed by a dedicated
  chain link.
  \item We present \textbf{SkillChain}, a deployed image-based framework that
  closes all three Skill feedback loops in a single self-evolving E-commerce
  lifecycle with a stage-wise monotone quality guarantee.
  \item Production validation across five visual intent categories confirms
  strictly additive stage gains, with online A/B evidence over the deployed baseline.
\end{enumerate}

\section{Related Work}

LLM-based autonomous agents have seen rapid progress across diverse
domains~\citep{wang2024agent_survey,xi2023rise}; we organize related
work into four threads most pertinent to SkillChain.

\paragraph{Skill lifecycle and self-evolving agent systems.}
Voyager~\citep{wang2023voyager} and Ghost in the Minecraft~\citep{zhu2023ghost}
pioneered reusable code-block and sub-goal skill libraries for open-world
exploration.
Building on this, AutoSkill~\citep{yang2026autoskillexperiencedrivenlifelonglearning},
SkillForge~\citep{liu2026skillforge}, SkillClaw~\citep{ma2026skillclaw},
CoEvoSkills~\citep{zhang2026coevoskillsselfevolvingagentskills}, and
EvoSkill~\citep{alzubi2026evoskillautomatedskilldiscovery} evolve skills
from interaction traces via failure-driven refinement, trace aggregation, or
co-evolved surrogate verifiers;
Trace2Skill~\citep{ni2026trace2skilldistilltrajectorylocallessons},
XSkill~\citep{jiang2026xskillcontinuallearningexperience}, and
WebXSkill~\citep{wang2026webxskillskilllearningautonomous} distill trajectory pools;
SkillRL~\citep{xia2026skillrl}, ARISE~\citep{li2026ariseagentreasoningintrinsic},
and SkillOS~\citep{ouyang2026skilloslearningskillcuration} apply RL-based curation;
SkillX~\citep{wang2026skillx}, Graph of Skills~\citep{liu2026graphskillsdependencyawarestructural},
and SkillNet~\citep{liang2026skillnetcreateevaluateconnect} construct structured
skill repositories; the Dual-Granularity Skill Bank~\citep{tu2026dynamicdualgranularityskillbank}
maintains coarse-to-fine skill abstractions for agentic RL.
\citet{jiang2026sok} and \citet{sumers2023cognitive} argue that agent skills
constitute a capability class distinct from tool use, generalizing across tasks
and modalities.
SkillChain differs in three key respects:
(1)~Skills are \emph{declarative} behavioral specifications rather than executable code or memory;
(2)~it adds an explicit \emph{routing optimization} stage absent from all prior work;
(3)~it is validated at industrial e-commerce scale.

\paragraph{Tool-augmented LLMs.}
ReAct~\citep{yao2023react} and Reflexion~\citep{shinn2023reflexion} established
reasoning-action interleaving and verbal self-improvement for tool use;
ToolFormer~\citep{schick2023toolformer} and Gorilla~\citep{patil2023gorilla}
further extended LLM tool use to large API libraries.
SkillChain instead manages \emph{specifications} governing tool invocation and
continuously refines them from production feedback.

\paragraph{Automatic prompt optimization.}
OPRO~\citep{yang2023opro}, APE~\citep{zhou2023ape}, DSPy~\citep{khattab2023dspy},
and TextGrad~\citep{yuksekgonul2024textgrad} optimize prompts via LLM feedback
or text-based differentiation;
ExpeL~\citep{zhao2024expel} distills execution traces into reusable templates.
SkillChain shares the data-driven improvement motivation but operates over
structured, intent-specific Skill Bodies with production routing signals.

\paragraph{LLM-as-Judge evaluation.}
G-Eval~\citep{liu2023geval}, MT-Bench~\citep{zheng2023judging}, Constitutional
AI~\citep{bai2022constitutional}, and HELM~\citep{liang2022helm} establish
LLM-based quality assessment, AI-driven constraint enforcement, and holistic
multi-metric evaluation, all informing our four-dimensional scoring design.
We adopt the LLM-as-Judge paradigm but ground evaluation in Skill Body
constraints, enabling feedback directly actionable for Body refinement.

\begin{figure*}[t]
	\centering
	\includegraphics[width=160mm]{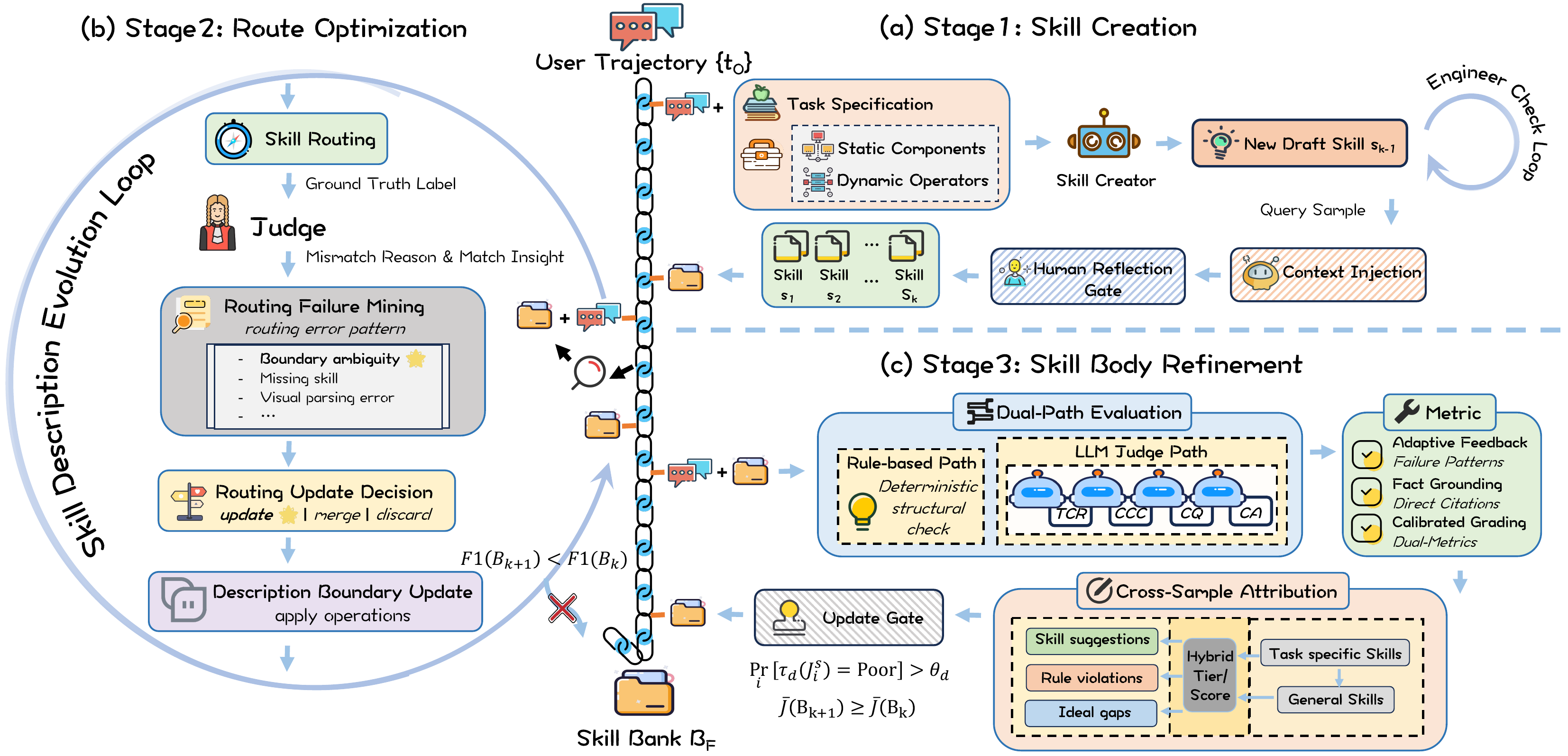}
	\caption{An overview of the SkillChain three-stage framework.
    \textbf{(a) Skill Creator} bootstraps Skills from task specifications
    and user trajectories via an engineer loop, then gates quality through
    human reflection before deployment.
    \textbf{(b) Route Optimizer} continuously mines routing failures in production and applies update/merge/discard operations to keep Skill
    Description boundaries aligned with real traffic.
    \textbf{(c) Body Refiner} evaluates responses through a dual-path
    pipeline and aggregates cross-sample
    signals to drive iterative Skill Body refinement.}
 \label{fig:overview}
\end{figure*}

\section{Methodology}
\label{sec:methodology}

Production Skills degrade along three independent dimensions:
initialization quality, routing accuracy, and behavioral compliance.
Because the Description ($d$) governing routing and the Body ($b$) governing
response quality are architecturally decoupled, their feedback loops close
sequentially without interference, embodying the \emph{unidirectional chain} design
of SkillChain.

Following \citet{jiang2026sok}, a \textbf{Skill} is a tuple
$s = (d,\ b,\ C_s,\ O_d)$: $d$ governs routing, $b$ specifies format, tool,
and constraint rules, $C_s$ provides static knowledge and examples, and $O_d$
lists dynamic operators invoked at inference time.
All Skills are versioned in a \emph{Skill Bank} $\mathcal{B}_k$; SkillChain
automates the full lifecycle across three coupled stages,
as illustrated in Figure~\ref{fig:overview}.

\subsection{Stage 1: Skill Creation}

\paragraph{LLM-Driven Bootstrap.}
Stage~1 bootstraps a Skill from a Task Specification, User Trajectories, i.e.,
real interaction sequences characterizing the target intent, and the current Skill
Bank $\mathcal{B}_k$ for knowledge reuse.
An LLM acting as Skill Creator generates an initial draft conditioned on these
inputs, using reference Skills retrieved from $\mathcal{B}_k$~\citep{lewis2020rag}
as in-context exemplars.
The draft is then refined through an \emph{Engineer Loop}~\citep{wu2024autogen}
that validates Static Components and Dynamic Operators against sampled queries, terminating
when all declarations pass correctness checks.

\paragraph{Human Reflection Gate.}
Before deployment, the Optimized Skill is tested against a curated query
sample and reviewed by domain experts for Body content
quality~\citep{ouyang2022instructgpt}; only approved Skills are versioned
into $\mathcal{B}_{k+1}$.
This gate catches Body-level deficiencies that programmatic checks cannot
capture, but it cannot anticipate routing drift under the full production
traffic distribution, the coverage problem that Stage~2 specifically targets.
Different intent types impose structurally distinct Body constraints;
Table~\ref{tab:intent} summarizes these per-intent requirements.

\begin{table}[t]
  \centering
  \small
  \begin{tabular}{lp{4.2cm}}
    \toprule
    \textbf{Intent} & \textbf{Key Constraint} \\
    \midrule
    Exact Match        & Product search via image or text query; concise card output \\
    Multi-Product      & Multi-item scene decomposition; per-item structured comparison \\
    Divergent Rec.     & Style-based recommendation; copy extends from image \\
    Encyclopedia       & Visual subject identification; knowledge response \\
    Utility Assistance & Scene task via dedicated tools; grounded output, no fabrication \\
    \bottomrule
  \end{tabular}
  \caption{Scene-specific Skill design principles applied during Stage~1.}
  \label{tab:intent}
\end{table}

\subsection{Stage 2: Route Optimization}

As traffic evolves, Skill Descriptions that were precise at creation time
drift out of alignment with user intents matched by the backbone
MLLM~\citep{qwen3vl2025}.

\paragraph{Routing Failure Analysis.}
To detect and repair this drift, a Judge LLM compares sampled routing decisions
against human-annotated ground truth, collecting failures into a routing
failure pool.
An LLM classifier then assigns each failure one of three root-cause labels
(Appendix~\ref{sec:routing-prompt}):
\emph{Boundary ambiguity}, where the Description does not clearly include or
exclude the query; \emph{Missing Skill}, where no existing Skill covers the
intent; or \emph{Visual parsing error}, which is escalated to the upstream
parser.
In our deployment, boundary ambiguity accounts for the large majority of
failures; once the initial Skill Bank covers core intent categories, missing
Skill cases are rare, confirming that Description boundary maintenance is the
dominant routing cost.

\paragraph{Iterative Description Update.}
Stage~2 runs iteratively over a held-out validation set, accepting an update only when
$\mathrm{F1}(\mathcal{B}_{k+1}) \geq \mathrm{F1}(\mathcal{B}_k)$,
which provides a monotone quality guarantee.
Based on mined failure patterns, each round applies one of three operations to
$\mathcal{B}_k$~\citep{yang2026autoskillexperiencedrivenlifelonglearning}:
\begin{align}
\text{\textbf{Update}}: \quad & \mathcal{B}_{k+1} = \mathcal{B}_k \text{ with } d_s \leftarrow d_s' \notag\\
\text{\textbf{Merge}}: \quad & \mathcal{B}_{k+1} = (\mathcal{B}_k \setminus \{s_i, s_j\}) \cup \{s'\} \notag\\
\text{\textbf{Discard}}: \quad & \mathcal{B}_{k+1} = \mathcal{B}_k \setminus \{s\} \label{eq:stage2}
\end{align}

\subsection{Stage 3: Skill Body Refinement}
\label{sec:stage3}

Even after routing is repaired, Skill Bodies accumulate deficiencies that
only surface at production scale: format violations, content gaps, tool
misuse.
Individual-sample feedback is too noisy to act on: per-query variation in
content and phrasing causes high variance in any single Judge evaluation,
and acting on individual scores risks overcorrecting to idiosyncratic cases
rather than systematic weaknesses.

\paragraph{Dual-Path Evaluation.}
A \emph{dual-path evaluation} pipeline covers complementary failure modes:
a rule-based path for deterministic structural checks and an LLM Judge
path~\citep{liu2023geval,zheng2023judging} that scores each response on four
dimensions: Tool Call Rationality (TCR), Card Composition Compliance (CCC),
Content Quality (CQ), and Constraint Adherence (CA), with
natural-language rationales grounding feedback in Skill Body constraints
(Appendix~\ref{sec:judge-prompt}).
Neither path alone captures the full failure space.

\paragraph{Cross-Sample Attribution.}
Rather than acting on individual scores, Stage~3 discretizes each Judge score
into three tiers (Good / Average / Poor) and computes tier distributions per
Skill across all attributed responses; a Skill is flagged for dimension $d$
when
\begin{equation}
\Pr_{i}\!\left[\tau_d(J_i^s) = \text{Poor}\right] > \theta_d
\label{eq:flag}
\end{equation}
where $\tau_d(\cdot)$ maps a score to its tier and $\theta_d$ is a
dimension-specific threshold calibrated empirically to reflect each dimension's
natural score variance and the minimum signal-to-noise ratio needed to
distinguish genuine Skill-level weaknesses from per-query fluctuations
(see Appendix~\ref{sec:impl-details}).
An LLM then synthesizes these distributions and per-sample rationales into
structured directives: Skill Suggestions, Rule Violations, and Ideal
Gaps~\citep{yuksekgonul2024textgrad,zhao2024expel}, targeting recurring
deficiencies rather than one-off failures.

\paragraph{Update Gate.}
Refined Skills re-enter the Human Reflection gate before deployment, and
Body edits are accepted only when $\bar{J}(\mathcal{B}_{k+1}) \geq \bar{J}(\mathcal{B}_k)$,
completing the monotone quality guarantee.
Because each chain link modifies a disjoint component, where Stage~2 only updates
$d$ and Stage~3 only updates $b$, the two stages can run over the same live
Skill Bank in either sequential or alternating fashion without invalidating
each other's updates.

\section{Experiments}
\label{sec:exp}

\subsection{Experimental Setup}
\label{sec:setup}

\paragraph{Systems.}
We compare five cumulative configurations: \textbf{NoSkill} (production
baseline without Skill constraints), \textbf{ManualSkill} (human-crafted
Skills, no SkillChain pipeline), and three SkillChain variants each adding
one stage to the previous: \textbf{Stage~1} (Skill creation),
\textbf{Stage~2} ($+$routing optimization), and \textbf{Stage~3}
($+$Body refinement, full SkillChain).
ManualSkill and SkillChain use the same backbone MLLM, tool inventory, card
schema, response protocol, task specifications, representative examples, and
knowledge sources. ManualSkill is therefore a static human-authored
specification baseline, while SkillChain starts from the same resources and
adds structured production feedback through routing-failure mining and
cross-sample Judge attribution.
Full system descriptions are in Appendix~\ref{sec:impl-details}.

\paragraph{Evaluation set.}
The offline set comprises 1{,}000 production queries sampled with
intent-stratified sampling across all five intent categories
listed in Table~\ref{tab:intent}.
The online A/B experiment runs for one week comparing Stage~3 against
the deployed Stage~2 baseline, followed by an additional 7-day cohort
tracking window for the complete return-rate metric; the deployment rationale
is discussed in Appendix~\ref{sec:impl-details}.

\paragraph{Metrics.}
\label{sec:metrics}
\textit{Offline}: four LLM-Judge dimensions~\citep{liu2023geval,zheng2023judging} (Tool
Call Rationality (TCR), Card Composition Compliance (CCC), Content Quality
(CQ), and Constraint Adherence (CA), each $\in [0,100]$ after score
normalization) and Routing Accuracy F1 against human-annotated ground
truth~\citep{liang2022helm}.
\textit{Online}: Interactive UV, Full-read Rate, Avg.\ Dwell Time, and
Return Rate.
Formal definitions and scoring rubrics are in Appendix~\ref{sec:metric-defs}.

\subsection{Main Results}
\label{sec:main}

\begin{table*}[!tbp]
  \centering
  \small
  \setlength{\tabcolsep}{5pt}
  \begin{tabular}{lp{6.0cm}cccccc}
    \toprule
    \multirow{2}{*}{\textbf{Method}}
      & \multirow{2}{*}{\textbf{Added Component}}
      & \multicolumn{5}{c}{\textbf{LLM Judge Dimensions} ($\uparrow$)}
      & \textbf{Routing} \\
    \cmidrule(lr){3-7}
      & & \textbf{TCR} & \textbf{CCC}$^\dagger$ & \textbf{CQ} & \textbf{CA}
      & \textbf{Avg}
      & \textbf{F1} ($\uparrow$) \\
    \midrule
    NoSkill
      & Production baseline (no Skill constraint)
      & 54.4 & 49.1 & 69.6 & 52.8 & 59.1 & --- \\
    ManualSkill
      & Human-designed skill baseline
      & \cellcolor{bestcol}\textbf{61.4}
      & 56.1
      & 70.9
      & \cellcolor{bestcol}\textbf{65.3}
      & 64.9
      & 61.5 \\
    \midrule
    SkillChain$_{S1}$
      & Auto Skill creation + human reflection
      & 46.7
      & 58.9
      & \cellcolor{secondcol}\underline{76.6}
      & 53.7
      & 62.5
      & \cellcolor{secondcol}\underline{65.5} \\
    SkillChain$_{S1+S2}$
      & $+$ Routing failure mining \& Description update
      & 55.3
      & \cellcolor{secondcol}\underline{63.4}
      & 76.5
      & 54.1
      & \cellcolor{secondcol}\underline{65.2}
      & \cellcolor{bestcol}\textbf{78.0} \\
    \textbf{SkillChain} (Full)
      & $+$ Dual-path evaluation \& Body refinement
      & \cellcolor{secondcol}\underline{61.3}
      & \cellcolor{bestcol}\textbf{72.2}
      & \cellcolor{bestcol}\textbf{82.3}
      & \cellcolor{secondcol}\underline{62.7}
      & \cellcolor{bestcol}\textbf{72.2}
      & \cellcolor{bestcol}\textbf{78.0} \\
    \bottomrule
  \end{tabular}
  \caption{
    Main offline evaluation results.
    Each SkillChain row adds one pipeline stage cumulatively over the previous.
    LLM Judge dimensions are each scored in $[0,100]$:
    \textbf{TCR} Tool Call Rationality,
    \textbf{CCC} Card Composition Compliance,
    \textbf{CQ} Content Quality,
    \textbf{CA} Constraint Adherence;
    \textbf{Avg} normalized aggregate score across all four dimensions.
    $^\dagger$CCC computed on the subset of queries with product card output.
    \colorbox{bestcol}{\textbf{Bold}}: best per column;
    \colorbox{secondcol}{\underline{underline}}: second best.
  }
  \label{tab:main}
\end{table*}

Table~\ref{tab:main} reports offline LLM Judge scores and Routing F1 for all
five system configurations, with each SkillChain row adding one stage
cumulatively. Each stage addresses one challenge and contributes in the expected direction.

\paragraph{Stage~1.}
Stage~1 establishes a strong baseline by introducing per-intent behavioral
constraints (C1), improving three of four Judge dimensions over NoSkill
(CCC, CQ, and CA).
The TCR decline is mechanistically expected: without Skill constraints,
NoSkill's tool calls are governed by the model's general priors;
Stage~1 imposes per-intent tool-call specifications that improve TCR only
when routing is correct, but misrouted queries receive tool mandates
misaligned with their actual intent.
Unlike unconstrained generation, which at worst omits the optimal
tool, an incorrect Skill Body \emph{actively forces} erroneous calls,
pushing aggregate TCR below the NoSkill baseline.

\paragraph{Stage~2.}
Stage~2 directly targets routing drift (C2), with gains concentrated in
Routing F1 and TCR.
Routing repair reduces wrong-constraint exposure, restoring TCR once
the correct skill is consistently selected; accurate intent-to-Skill
mapping is a prerequisite for correct tool invocation.

\paragraph{Stage~3.}
Stage~3 closes the quality feedback loop (C3), with the largest incremental
gains in CCC and CA.
The rule-based path directly targets structural violations captured by CCC, while the LLM Judge's CA dimension checks
Skill Body compliance, making both the primary beneficiaries of the Stage~3
loop.

\paragraph{vs.\ ManualSkill.}
Stage~3 achieves the highest aggregate score and outperforms ManualSkill on
CCC, CQ, and Routing F1; the CA gap reflects the denser constraint
specification of self-evolved Skills, which raises the bar for full adherence
(see Appendix~\ref{sec:skill-comparison}).
Because CCC is evaluated only when a system emits product cards, we also
report the card-emission denominator in Table~\ref{tab:card-emission}.
Full SkillChain has the highest card-emission rate and a comparable average
card count to the Skill baselines, so its CCC gain is not caused by
selectively emitting fewer cards.

\begin{table}[!tbp]
  \centering
  \small
  \begin{tabular}{lcc}
    \toprule
    \textbf{System} & \textbf{Avg.\ card count} & \textbf{Emission rate} \\
    \midrule
    NoSkill & 2.05 & 64.9\% \\
    ManualSkill & 3.69 & 90.8\% \\
    SkillChain$_{S1}$ & 3.67 & 99.0\% \\
    SkillChain$_{S1+S2}$ & 3.33 & 96.5\% \\
    \textbf{SkillChain} (Full) & 3.62 & 99.5\% \\
    \bottomrule
  \end{tabular}
  \caption{
    Card-emission denominator for CCC. Full SkillChain has the highest
    card-emission rate and a comparable average card count to the Skill
    baselines, so its CCC gain is not caused by selectively emitting fewer
    cards.
  }
  \label{tab:card-emission}
\end{table}

\subsection{Analysis}
\label{sec:analysis}

\begin{table}[!tbp]
  \centering
  \small
  \setlength{\tabcolsep}{5pt}
  \resizebox{\linewidth}{!}{%
  \begin{tabular}{lcccccc}
    \toprule
    \multirow{2}{*}{\textbf{Intent}} &
      \multicolumn{1}{c}{\textbf{Stage 2}} &
      \multicolumn{5}{c}{\textbf{Stage 3}} \\
    \cmidrule(lr){2-2}\cmidrule(lr){3-7}
    & \textbf{F1} $\Delta$
    & \textbf{TCR} $\Delta$ & \textbf{CCC} $\Delta$  & \textbf{CQ} $\Delta$ & \textbf{CA} $\Delta$ & \textbf{Avg} $\Delta$\\
    \midrule
    Exact Match        &$+$3.1 &$+$6.4 &$+$8.0 &$+$2.8 &$+$2.3 &$+$4.4 \\
    Multi-Product      &$+$11.3 &$+$6.8 &$+$3.5 &$+$2.4 &$+$1.1 &$+$3.2 \\
    Divergent Rec.     &$+$18.0 &$+$7.1 &$+$12.7 &$+$1.4 &$+$3.1 &$+$5.2 \\
    Encyclopedia       &\textbf{$+$18.5} &\textbf{$+$13.3} &\textbf{$+$15.6} &\textbf{$+$3.7} &\textbf{$+$5.8} &\textbf{$+$8.4} \\
    Utility Assistance &$+$9.7 &$-$3.5 &$+$7.1 &$-$0.1 &$-$2.2 &$+$0.3 \\
    \bottomrule
  \end{tabular}}
  \caption{Per-intent incremental gains from Stage~2 (routing, $\Delta$ over Stage~1) and Stage~3 (body quality, $\Delta$ over Stage~2) across five intent categories. \textbf{Bold}: best per column.}
  \label{tab:per_intent}
\end{table}

We design three research questions to probe SkillChain's core design claims:
(Q1)~whether routing and response quality gains follow the unidirectional
chain order without reverse propagation;
(Q2)~whether the self-evolution converges and at what rate;
(Q3)~what drives Stage~3 quality gains.

\paragraph{RQ1: Chain Unidirectionality.}
Table~\ref{tab:main} confirms the unidirectionality hypothesis: Stage~2 gains
concentrate on Routing F1 and TCR with negligible changes to CQ and CA;
Stage~3 reverses the pattern, with the largest gains in CCC and CA and a
matched Routing F1.
This dissociation reflects the unidirectional chain design: gains accumulate
strictly downstream without reversing upstream metrics.

Table~\ref{tab:per_intent} breaks down per-intent gains within each stage and
reinforces this downstream dependency: Stage~2 gains are largest for
Encyclopedia and Divergent Rec., where routing drift was most pronounced, and
Stage~3's largest Body-quality gains occur on Encyclopedia, where substantial
content and card-composition headroom remained after routing repair. Utility
Assistance shows near-zero Stage~3 gains, reflecting the difficulty of
exhaustively constraining open-ended intents.

\begin{figure}[t]
    \centering 
    \includegraphics[scale=0.42]{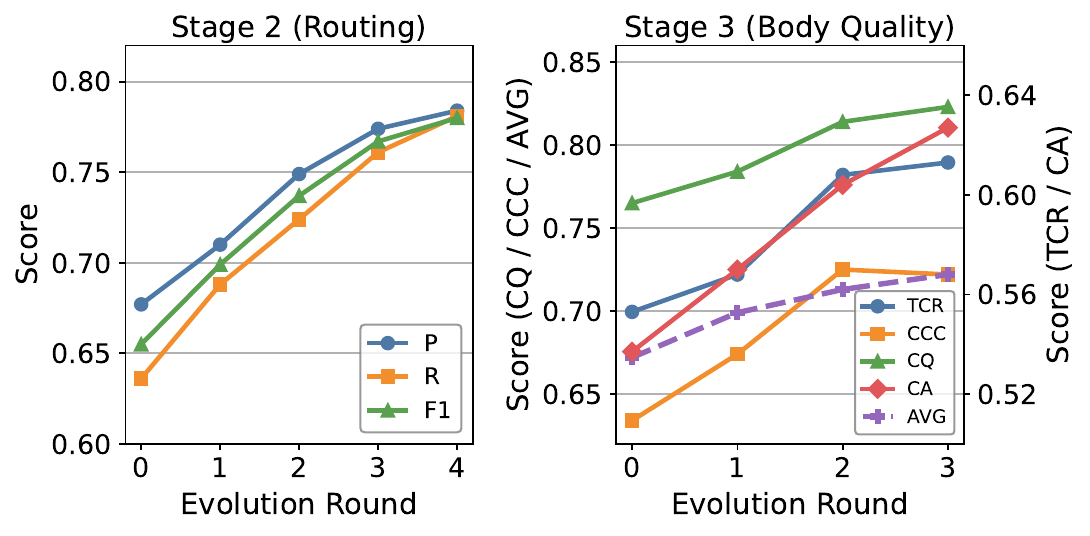}
    \caption{Evolution of routing quality (Stage~2 Routing P, R, F1) and body quality metrics (CCC, TCR, CQ, CA) across optimization rounds. Stage~2 converges within four rounds; structural dimensions (CCC, TCR) plateau earlier than content-oriented ones (CQ, CA). Two stage trajectories evolve independently, confirming that downstream refinement does not reverse upstream routing gains.}
    \label{fig:evolution}
\end{figure}

\paragraph{RQ2: Evolution Convergence.}
Figure~\ref{fig:evolution} tracks how routing quality and Body quality evolve
across optimization rounds.
Stage~2 Routing F1 rises steadily over four rounds before plateauing, with
boundary Update operations contributing the most.
Stage~3 exhibits dimension-level heterogeneity: structural dimensions (CCC,
TCR) plateau earlier while content-oriented dimensions (CQ, CA) continue
improving across all three rounds.
Structural violations are finite and discrete: a small set of rule-based
signals can enumerate and repair them within a few rounds, whereas
content-quality failures span a long tail of query phrasings and domain
sub-topics, requiring more production exposure to converge.
The two trajectories evolve independently, empirically confirming that the
unidirectional chain design holds at runtime.

\paragraph{RQ3: Stage 3 Design Choices.}
\label{sec:rq3}

Table~\ref{tab:ablation-stage3} ablates the two key design choices in Stage~3.
First, the two evaluation paths are necessary but complementary.
Removing the LLM Judge path incurs the larger overall penalty, with TCR and
CA dropping the most, as both metrics require holistic understanding of task
intent and are difficult to evaluate with fixed rules.
Removing the rule-based path causes a distinct pattern: CCC suffers the most
while CQ is almost unaffected, confirming that rule-based signals capture
structural violations an LLM Judge tends to miss.
Second, Statistical Aggregation is the more critical attribution component.
Removing it collapses CCC by $-12.2$, the largest single-dimension drop across
all ablations, and lowers Avg by $-3.8$, indicating that without cross-sample
aggregation the system cannot distinguish per-query noise from genuine
Skill-level weaknesses.
Removing Qualitative Induction has a much milder effect on Avg, and CA even
improves marginally, suggesting that qualitative inductive summaries
occasionally introduce spurious rewrites for content-oriented metrics.

\begin{table*}[!tbp]
  \centering
  \footnotesize
  \setlength{\tabcolsep}{4pt}
  \begin{tabular}{lccccccccccc}
    \toprule
    \multirow{2}{*}{\textbf{Variant}}
      & \multicolumn{2}{c}{\textbf{TCR}}
      & \multicolumn{2}{c}{\textbf{CCC}}
      & \multicolumn{2}{c}{\textbf{CQ}}
      & \multicolumn{2}{c}{\textbf{CA}}
      & \multicolumn{2}{c}{\textbf{Avg}} \\
    \cmidrule(lr){2-3}\cmidrule(lr){4-5}\cmidrule(lr){6-7}\cmidrule(lr){8-9}\cmidrule(lr){10-11}
      & Score & $\Delta$ & Score & $\Delta$ & Score & $\Delta$ & Score & $\Delta$ & Score & $\Delta$ \\
    \midrule
    \multicolumn{11}{l}{\textit{(a) Evaluation signal source}} \\
    Dual-path (full)                        & 61.3 & \phantom{$-$}-- & 72.2 & \phantom{$-$}-- & 82.3 & \phantom{$-$}-- & 62.7 & \phantom{$-$}-- & 72.2 & \phantom{$-$}-- \\
    \quad $-$ \textit{LLM Judge path}       & 55.7 & $-$5.6 & 69.2 & $-$3.0 & 78.9 & $-$3.4 & 57.3 & $-$5.4 & 68.0 & $-$4.2 \\
    \quad $-$ \textit{Rule-based path}      & 58.4 & $-$2.9 & 68.4 & $-$3.8 & 81.9 & $-$0.4 & 60.8 & $-$1.9 & 70.3 & $-$1.9 \\
    \midrule
    \multicolumn{11}{l}{\textit{(b) Cross-sample attribution level}} \\
    Skill-level $+$ Induction (full)        & 61.3 & \phantom{$-$}-- & 72.2 & \phantom{$-$}-- & 82.3 & \phantom{$-$}-- & 62.7 & \phantom{$-$}-- & 72.2 & \phantom{$-$}-- \\
    \quad $-$ \textit{Qualitative Induction}   & 59.0 & $-$2.3 & 70.9 & $-$1.3 & 81.7 & $-$0.5 & 63.0 & $+$0.3 & 71.3 & $-$0.9 \\
    \quad $-$ \textit{Statistical Aggregation} & 57.6 & $-$3.7 & 60.0 & $-$12.2 & 81.4 & $-$0.9 & 61.5 & $-$1.2 & 68.4 & $-$3.8 \\
    \bottomrule
  \end{tabular}
  \caption{Ablation on Stage~3 design choices.
  $\Delta$ reports the change relative to the full variant within each group
  (\phantom{$-$}-- denotes the reference row).
  Metrics defined in \S\ref{sec:metrics}.}
  \label{tab:ablation-stage3}
\end{table*}

\subsection{Online A/B Experiment}
\label{sec:ab}

\begin{table}[!tbp]
  \centering
  \small
  \begin{tabular}{lc}
    \toprule
    \textbf{Metric} & \textbf{$\Delta$ (Stage~3 vs.\ Stage~2)} \\
    \midrule
    Interactive UV       & $+$1.92 pp \\
    Full-read Rate       & $+$4.98 pp \\
    Avg.\ Dwell Time (s) & $+$2.85    \\
    Same-week Return Rate & $+$1.15 pp \\
    Complete 7-day Return Rate & $+$1.61 pp \\
    \bottomrule
  \end{tabular}
  \caption{
    Online A/B results (Stage~3 vs.\ Stage~2, one-week production experiment).
    $\Delta$ values are absolute improvements (pp\,=\,percentage points).
    Interactive UV counts unique users who either ask follow-up questions
    (knowledge queries) or click a product card (commercial queries).
    Same-week Return Rate is measured within the experiment week, while
    Complete 7-day Return Rate follows the exposed cohort for another 7 days.
  }
  \label{tab:online}
\end{table}

Table~\ref{tab:online} reports online metric deltas for Stage~3 vs.\ the
already-deployed Stage~2 baseline over a one-week production A/B experiment.
The test bucket contains around 80K daily active users under the production
traffic allocation, and all proportion metrics are significant under a
two-sided two-proportion z-test ($p<0.001$).
Full-read Rate shows the largest gain ($+$4.98\,pp) and same-week
within-experiment Return Rate improves by $+$1.15\,pp, signalling sustained
attention beyond single-session effects.
Because a complete 7-day return window cannot be fully observed inside a
one-week experiment, we additionally followed the exposed cohort for another
7 days after the experiment; the complete 7-day Return Rate gain is
$+$1.61\,pp.
Interactive UV ($+$1.92\,pp) and Dwell Time ($+$2.85\,s) confirm gains across
both commercial and knowledge-seeking queries, consistent with the offline
LLM-as-Judge quality improvements.

\subsection{Human Evaluation (SBS)}
\label{sec:sbs}

\begin{figure}[t]
    \centering
    \includegraphics[scale=0.5]{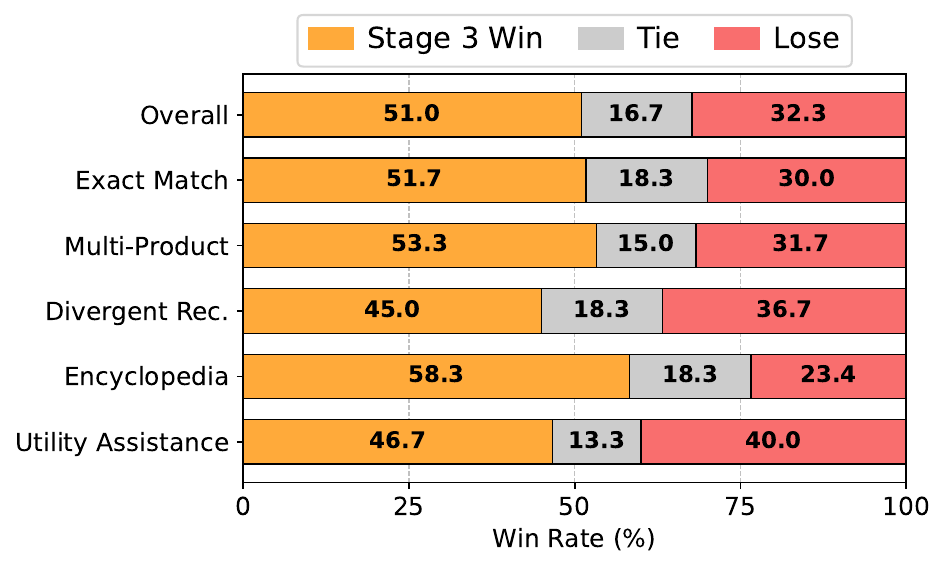}
    \caption{SBS human evaluation win/tie/lose rates for Stage~3 vs.\ ManualSkill across five intent categories (300 queries, 60 per intent). SkillChain achieves a net win across all intents, with the largest margin on content-demanding scenes.}
    \label{fig:winrates}
\end{figure}

We conduct a side-by-side (SBS) blind evaluation with human annotators
comparing Stage~3 vs.\ ManualSkill responses on a balanced set of 300 queries
(60 per intent category).
Annotators see paired responses in randomized order, with system identifiers
and other identifying cues removed, and assign win/tie/lose labels.
Following a double-check annotation protocol, each item is reviewed by
annotators; cases with inconsistent labels are adjudicated by a senior
annotator to obtain the final SBS label. The raw disagreement rate before
adjudication is 9\%, and the final adjudicated labels reach Cohen's
$\kappa=0.85$.
Figure~\ref{fig:winrates} reports win/tie/lose rates overall and per scene.
Stage~3 achieves a consistent net win across all five intent
categories.
Encyclopedia and Utility Assistance show the largest human preference margins,
while Divergent Rec.\ is the weakest, reflecting the subjective nature of
style and diversity preferences in recommendation responses.
Notably, Utility Assistance yields a strong human preference despite its
near-zero offline Stage~3 gains, suggesting that the LLM Judge underestimates
quality improvements for open-ended intents, consistent with the attribution
noise identified in \S\ref{sec:analysis}.

To validate the LLM Judge against SBS labels, we convert Judge tiers to
ordinal scores (Good=2, Average=1, Poor=0) and derive automatic win/lose
preferences by comparing the two systems per metric; the aggregate preference
uses TCR $+$ CCC $+$ 2CQ $+$ CA. After excluding tie cases, agreement is
strongest for CCC, CQ, and CA, while TCR remains noisier. Full agreement
statistics are reported in Appendix~\ref{sec:judge-validation}.

\section{Conclusion}
\label{sec:conclusion}

We presented SkillChain, a framework that closes the loop on Skill evolution
for e-commerce AI assistants.
By decomposing the problem into Skill creation (Stage~1), routing optimization
(Stage~2), and Body refinement (Stage~3), SkillChain continuously adapts to
real traffic distribution and quality signals without manual re-engineering.
Experiments on the production system demonstrate significant offline and
online improvements across diverse visual intent types.
The Skill abstraction is modality-agnostic; future work may extend it to
text-based search, live-stream commentary, and product Q\&A settings with
similar intent diversity.

\clearpage
\section{Limitations}

SkillChain has several limitations worth noting.
First, \textbf{cold-start and Skill Bank scalability}: intents absent from
the initial Skill Bank fall back to NoSkill, making coverage dependent on
upfront seeding breadth; routing disambiguation complexity grows with Skill
count, and routing boundary conditions between overlapping intents remain
inherently difficult to delineate.
Second, \textbf{evaluation reliability}: LLM Judge quality degrades for
open-ended intents where no single ground-truth response exists to anchor
scoring, which explains the near-zero Stage~3 gains for Utility Assistance, and
Stage~2 routing failure diagnosis relies on human-annotated ground-truth
routing labels, limiting applicability where such supervision is scarce.
Third, \textbf{declarative representation ceiling and generalizability}:
Skills are flat text specifications and cannot encode executable procedures or
hierarchical sub-skill compositions, capping expressiveness for complex
multi-step tasks; all experiments are conducted on a single e-commerce
platform with five intent categories, and generalizability to other domains
and languages remains to be validated.
We plan to address these through automatic Skill discovery,
multi-judge consensus scoring, executable Skill representations,
and cross-platform evaluation in future work.

\bibliography{custom}

\appendix
\clearpage

\section{Human-Crafted vs.\ Self-Evolved Skill Comparison}
\label{sec:skill-comparison}

\paragraph{Representative case.}
Table~\ref{tab:skill_comparison} illustrates how SkillChain's self-evolution
produces qualitatively different Skills compared to manually authored ones,
using the Utility Assistance intent as a representative case.
This intent exemplifies a broader pattern: each intent type may encompass
multiple specialized Skills targeting distinct sub-scenes, and Utility
Assistance alone spans diverse sub-scenes, e.g., health consultation, recipe guidance, and document
assistance, each with its own Body constraints.

\paragraph{Structural differences.}
Human-crafted Skills capture high-level intent but leave routing boundaries
vague, tool invocation underspecified, and domain constraints generic.
SkillChain iteratively tightens each dimension through production
signals: routing failures refine Description boundaries (Stage~2), while
LLM Judge attribution identifies and repairs specific Body weaknesses
(Stage~3).
Beyond these three dimensions, self-evolved Skills also enumerate edge cases
and follow-up patterns that manual authoring leaves unaddressed, resulting in
roughly twice the constraint density (see Table~\ref{tab:skill_comparison}).

\paragraph{Metric interpretation.}
The increased constraint density also provides a lens for interpreting the
metric trade-offs observed in Table~\ref{tab:main}.
Self-evolved Skills carry roughly twice the specification volume of
ManualSkills, including mandatory field-level rules, banned-word blacklists,
strict invocation ordering, and enumerated edge-case handlers, making full
Constraint Adherence (CA) harder to achieve by construction: even a
well-formed response can now fail on a clause that ManualSkill never expressed.
This explains why CA in SkillChain Full ($62.7$) falls below the ManualSkill
baseline ($65.3$) despite a higher overall Avg.
From a user-facing perspective, however, CCC and CQ are the dimensions that
matter most: CCC reflects whether product cards are correctly composed
(directly visible to the user), and CQ captures overall content coherence.
Both improve substantially: CCC from $56.1$ to $72.2$ and CQ from $70.9$
to $82.3$, confirming that the richer constraint set drives meaningful
quality gains on the dimensions users actually experience.

\begin{table*}[h]
  \centering
  \small
  \setlength{\tabcolsep}{6pt}
  \begin{tabular}{@{}lp{5cm}p{6.5cm}@{}}
    \toprule
    \textbf{Aspect} & \textbf{ManualSkill} & \textbf{SkillChain (Self-Evolved)} \\
    \midrule
    Routing boundary
      & Intent described by name and high-level goal; no disambiguation from adjacent intents by default
      & Adds an explicit \emph{routing-fallback} block listing 3--5 named excluded scenarios with prescribed fallback targets  \\[6pt]
    Tool invocation
      & Available tools listed as a flat inventory; trigger conditions and call order unspecified; all search paths executed regardless of user scene
      & Selects the single most relevant search path based on user scene; enforces strict invocation order; ignores other branches \\[6pt]
    Format constraints
      & General structured-reply guidance; no field-level rules; card syntax informally described
      & Banned-word blacklists for Card \texttt{title}/\texttt{search} fields; mandatory closure; one-card-one-product rule; interleaving policy (inline text-then-card vs.\ end-clustered cards) specified per scenario; single-line JSON payload required \\[6pt]
    Domain constraints
      & Generic accuracy and helpfulness requirements; disclaimer usage at agent discretion
      & Vertical-specific rules derived from LLM Judge feedback: standardized \texttt{>}-blockquote disclaimer; mandatory emotional-close sentence before legal caveat; PII masking pattern; forbidden phrases paired with required replacement phrasings \\[6pt]
    Edge case handling
      & Not addressed; agent applies general reasoning to novel or boundary queries
      & 3--5 categorized follow-up question types per scenario; bans on generic closing phrases; boundary scenarios listed with explicit handling instructions \\[6pt]
    Constraint density
      & $\sim$60--80 specification lines per Skill
      & $\sim$150--200 specification lines per Skill ($\approx\!2\times$ increase) \\
    \bottomrule
  \end{tabular}
  \caption{Structural comparison of a ManualSkill vs.\ a SkillChain self-evolved Skill
  for the Utility Assistance intent across six dimensions.}
  \label{tab:skill_comparison}
\end{table*}

\section{Metric Definitions}
\label{sec:metric-defs}

\subsection{Offline LLM-Judge Dimensions}

The four dimensions form two complementary layers of response evaluation.
\textbf{TCR} and \textbf{CCC} assess \emph{structural execution fidelity}: whether
the response correctly follows the Dynamic Operators and rich-media formatting
prescribed by the Skill Body.
\textbf{CQ} and \textbf{CA} assess \emph{content compliance}: whether the
actual text meets domain quality standards and honors behavioral constraints.

Raw scores are summed (max 50) and linearly normalized to $[0,100]$:
\begin{equation}
\bar{J}(r,s) = 2\bigl(J_{\mathrm{TCR}} + J_{\mathrm{CCC}} + J_{\mathrm{CQ}} + J_{\mathrm{CA}}\bigr)
\label{eq:judge}
\end{equation}

\begin{itemize}
  \item \textbf{Tool Call Rationality (TCR)} [0--10] ($\uparrow$):
  targets the Dynamic Operators layer of the Skill Body.
  Measures whether tool selection and invocation order are appropriate for
  the query; penalizes redundant, missing, or sequentially incorrect calls.

  \item \textbf{Card Composition Compliance (CCC)} [0--10] ($\uparrow$):
  targets the rich-media composition rules and product quality in the Skill Body.
  Measures structural conformance of card output, and search term quality to the Skill specification.
  Evaluated only on the subset of queries that produce product card output.

  \item \textbf{Content Quality (CQ)} [0--20] ($\uparrow$):
  targets the writing quality and domain knowledge layer.
  Composite of factual accuracy [8\,pts], mobile-friendly readability
  [6\,pts], and intent understanding [6\,pts].
  In routing-deviation scenarios, intent understanding carries the most
  weight: rigid rule-following that misses the user's true need scores
  0--2 on this sub-dimension.

  \item \textbf{Constraint Adherence (CA)} [0--10] ($\uparrow$):
  targets the behavioral constraint layer, covering all mandatory rules and
  prohibitions explicitly enumerated in the Skill Body (e.g., no
  fabrication, format length limits, domain-specific disclaimers).
\end{itemize}

We additionally report \textbf{Routing Accuracy F1} ($\uparrow$), the
harmonic mean of routing precision and recall against human-annotated
ground-truth intent labels, following HELM~\citep{liang2022helm}:
\begin{equation}
F1_{\text{routing}} = \frac{2\,P_{\text{routing}}\cdot R_{\text{routing}}}{P_{\text{routing}} + R_{\text{routing}}}
\label{eq:f1}
\end{equation}
This metric is independent of the four Judge dimensions: it evaluates the
Description ($d$) layer independently of Body ($b$) compliance.

\subsection{Online A/B Metrics}

The four metrics cover three complementary levels of user behavior:
\emph{immediate engagement} (Interactive UV, Full-read Rate),
\emph{depth of attention} (Avg.\ Dwell Time), and \emph{cross-session
retention} (Return Rate).

\begin{itemize}
  \item \textbf{Interactive UV} ($\uparrow$): unique users who perform at
  least one active interaction within the session, such as clicking a product card
  or asking a follow-up question.

  \item \textbf{Full-read Rate} ($\uparrow$): proportion of sessions in
  which the user scrolls to the end of the response.
  Measures content consumption completeness.

  \item \textbf{Avg.\ Dwell Time} ($\uparrow$): mean time (seconds) spent
  on the response page per session.
  Complements Full-read Rate by distinguishing careful reading from fast
  scrolling.

  \item \textbf{Return Rate} ($\uparrow$): proportion of users who return to
  the assistant within the specified observation window, reported both as
  same-week within-experiment return and complete 7-day cohort return.
\end{itemize}

\section{LLM Judge Validation}
\label{sec:judge-validation}

Table~\ref{tab:judge-human-agreement} reports the full Judge-human agreement
statistics. Because near-zero score differences are difficult to align exactly
with human tie judgments, we exclude pairs where either the human or automatic
preference is Tie.

\begin{table}[!tbp]
  \centering
  \small
  \begin{tabular}{lcc}
    \toprule
    \textbf{Dimension} & \textbf{Exact agreement} & \textbf{Cohen's $\kappa$} \\
    \midrule
    TCR & 63.79\% & 0.32 \\
    CCC & 96.20\% & 0.88 \\
    CQ & 96.67\% & 0.91 \\
    CA & 90.54\% & 0.72 \\
    Avg & 82.35\% & 0.62 \\
    \bottomrule
  \end{tabular}
  \caption{
    Agreement between automatic LLM-Judge preferences and human SBS labels
    after excluding tie cases. Agreement is strongest for CCC, CQ, and CA,
    supporting the Judge for structured Body-quality diagnosis, while TCR is
    noisier.
  }
  \label{tab:judge-human-agreement}
\end{table}

\section{Implementation Details}
\label{sec:impl-details}

\paragraph{System configurations.}
Table~\ref{tab:impl} summarizes key implementation settings.
All system variants share the same backbone multimodal LLM,
\textbf{Qwen3-VL-235B-A22B-Instruct}~\citep{qwen3vl2025}, which is also
the production-deployed model.
The offline LLM Judge is \textbf{Gemini-3.1-Pro-Preview}~\citep{google2026gemini31},
prompted with the four-dimensional scoring rubric in Appendix~\ref{sec:judge-prompt};
all Judge calls use temperature~0 for reproducibility.
The Skill Creator in Stage~1 and the Skill Refiners in Stage~2 and Stage~3 are all powered by \textbf{Claude-Sonnet-4.6}~\citep{anthropic2026claude}.
The five experimental configurations are:
\begin{itemize}
  \item \textbf{NoSkill}: current production system; LLM generates responses
  without Skill constraints.
  \item \textbf{ManualSkill}: Skills hand-crafted by domain engineers without
  the SkillChain pipeline; serves as a static human-design baseline.
  \item \textbf{Stage~1} ($\text{Skill}_{v0}$): automatically generated Skills
  after human reflection; routing fixed, Body not refined.
  \item \textbf{Stage~2} ($+$Routing): Stage~1 Skills with Stage~2 routing
  optimization applied.
  \item \textbf{Stage~3} ($+$Refine): full SkillChain; Stage~2 with Stage~3
  Body refinement.
\end{itemize}

\paragraph{Online A/B deployment rationale.}
The online A/B experiment compares Stage~3 against the already-deployed
Stage~2 baseline rather than NoSkill, because Stage~2 had been launched
into large-scale production once its routing accuracy and Skill coverage met deployment
thresholds: correct intent routing is a prerequisite for acceptable user
experience at scale, so serving users without a stable routing layer was
not an option.
The online experiment therefore measures the incremental contribution of
Stage~3 Body refinement on top of a live, stable routing system.

\begin{table*}[h]
  \centering
  \small
  \setlength{\tabcolsep}{5pt}
  \begin{tabular}{llp{8cm}}
    \toprule
    \textbf{Component} & \textbf{Setting} & \textbf{Value} \\
    \midrule
    \multirow{2}{*}{Backbone MLLM}
      & Model    & Qwen3-VL-235B-A22B-Instruct \\
      & Usage    & Inference only (production-deployed; no fine-tuning) \\
    \midrule
    \multirow{3}{*}{LLM Judge}
      & Model       & Gemini-3.1-Pro-Preview \\
      & Temperature & 0 (deterministic) \\
      & Usage       & Offline evaluation only \\
    \midrule
    \multirow{2}{*}{Creator \& Refiner}
      & Model & Claude-Sonnet-4.6 \\
      & Usage & Stage~1 Skill creation; Stage~2 \& 3 Skill refinement \\
    \midrule
    \multirow{2}{*}{Routing Judge}
      & Source & Human-annotated ground-truth labels \\
      & Usage & Stage~2 failure classification \\
    \midrule
    \multirow{3}{*}{Stage~2}
      & Max iterations        & 4 \\
      & Convergence criterion & Routing F1 on held-out validation set \\
      & Failure pool per round & $\geq$30 samples per Skill \\
    \midrule
    \multirow{4}{*}{Stage~3}
      & Max rounds            & 3 \\
      & Poor-tier threshold   & TCR 20\,\%, CCC 10\,\%, CQ 5\,\%, CA 10\,\% \\
      & Min samples per Skill & 50 \\
    \midrule
    \multirow{2}{*}{Offline eval set}
      & Min per intent & 150 \\
      & Sampling       & Intent-stratified, proportional to traffic \\
    \midrule
    Online A/B
      & Duration & 1 week \\
    \bottomrule
  \end{tabular}
  \caption{Implementation settings for all SkillChain experiments.}
  \label{tab:impl}
\end{table*}

\section{Prompt Templates}
\label{sec:prompts}

\subsection{Routing Boundary Analysis Prompt}
\label{sec:routing-prompt}

The routing boundary analysis prompt drives Stage~2's Description refinement
loop by processing both misrouted cases (\textbf{Case A}) and correctly routed
cases (\textbf{Case B}).

In Case~A (mismatch), the model receives the image together with the predicted
routing, ground-truth routing, and Skill trigger boundary descriptions.
It identifies the specific visual features that caused the routing ambiguity,
explains the basis for the correct routing, and proposes a concrete one-sentence
update to the relevant Skill Description.
These suggestions are aggregated across failure cases to generate Description
refinement patches.

In Case~B (correct match), the model analyzes a correctly routed case to
reinforce existing boundary rules.
It identifies the visual features that positively support the routing, notes
any ambiguous elements and explains why the routing is still correct, and
summarizes what boundary rule the case reinforces.
This positive-evidence pass prevents over-tightening of Description boundaries
during refinement.
The full prompts for both cases are shown in Table~\ref{table:routing_prompt}.

\begin{table*}[!ht]
\small
\begin{tcolorbox}[
  enhanced,
  colback=gray!5, colframe=gray!40,
  colbacktitle=gray!55, coltitle=white,
  fonttitle=\small\bfseries, fontupper=\small,
  title={Routing Boundary Analysis Prompt Template},
  boxrule=0.8pt, arc=2pt,
  top=4pt, bottom=4pt, left=6pt, right=6pt
]

\textbf{\#\#\# Role}

You are a Skill routing boundary analysis expert.
Examine the input image carefully and analyze the routing case below.

\vspace{0.4em}
\noindent\rule{\linewidth}{0.4pt}
\vspace{-0.6em}

\textbf{Case A --- Mismatch (Routing Error)}

\textbf{Predicted:}~\textcolor[rgb]{0,0,0.7}{\{Predicted Skill\}} \quad
\textbf{Ground Truth:}~\textcolor[rgb]{0,0,0.7}{\{Label Skill\}}

\textbf{Skill Trigger Boundaries:}~\textcolor[rgb]{0,0,0.7}{\{Skill Descriptions\}}

\textbf{Analysis Tasks:}
\begin{enumerate}[leftmargin=1.5em,label=\arabic*.,topsep=1pt,itemsep=0pt]
  \item Describe the image: subject, background, quantity, text presence,
        product image vs.\ lifestyle shot, etc.
  \item Identify which visual features caused the routing ambiguity.
  \item Explain the basis for the correct routing.
  \item Propose a one-sentence Description update: name the Skill and
        which rule to add or tighten.
\end{enumerate}
\textbf{Output fields:}
\begin{itemize}[leftmargin=1.2em,topsep=1pt,itemsep=0pt]
  \item \texttt{image\_content}
  \item \texttt{error\_reason}
  \item \texttt{correct\_routing\_basis}
  \item \texttt{suggestion}
\end{itemize}

\vspace{0.4em}
\noindent\rule{\linewidth}{0.4pt}
\vspace{-0.6em}

\textbf{Case B --- Match (Correct Routing, Boundary Reinforcement)}

\textbf{Routing Result (Correct):}~\textcolor[rgb]{0,0,0.7}{\{Label Skill\}}

\textbf{Skill Trigger Boundaries:}~\textcolor[rgb]{0,0,0.7}{\{Skill Descriptions\}}

\textbf{Analysis Tasks:}
\begin{enumerate}[leftmargin=1.5em,label=\arabic*.,topsep=1pt,itemsep=0pt]
  \item Describe the image (same criteria as Case~A).
  \item Identify visual features that positively support this routing.
  \item Note any ambiguous elements and explain why routing here is still
        correct (\texttt{none} if not applicable).
  \item Summarize what boundary rule this case reinforces.
\end{enumerate}
\textbf{Output fields:}
\begin{itemize}[leftmargin=1.2em,topsep=1pt,itemsep=0pt]
  \item \texttt{image\_content}
  \item \texttt{routing\_evidence}
  \item \texttt{ambiguity\_risk}
  \item \texttt{boundary\_insight}
\end{itemize}

\end{tcolorbox}
\caption{Routing boundary analysis prompt templates used in Stage~2.
Case~A (mismatch) generates Description refinement suggestions;
Case~B (match) reinforces correct boundary rules.}
\label{table:routing_prompt}
\end{table*}

\subsection{LLM Judge Prompt Design}
\label{sec:judge-prompt}

The LLM Judge evaluates each response against the Skill Body constraints
across the four dimensions defined in Appendix~\ref{sec:metric-defs}.
A central design choice is the \textbf{Routing-Deviation Fallback Principle},
which overrides all other criteria: when the input does not perfectly match
the Skill's trigger scope, rigid rule-following, even if technically
compliant, is penalized to returning an error or silence.
The ideal response instead detects the deviation, infers the user's true
intent, and adaptively delivers a satisfying answer.
This prevents the Judge from rewarding safe-but-useless responses at Skill
boundaries.

The Judge produces structured JSON containing per-dimension scores and tiers
(Good / Average / Poor), \texttt{rule\_violations} and \texttt{ideal\_gaps}
(each requiring citation of concrete response excerpts), and
\texttt{skill\_md\_suggestions}.
The violations and gaps feed the cross-sample attribution step;
the suggestions are aggregated into Skill Body refinement patches.
The full prompt is shown in Table~\ref{table:judge_prompt}.

\begin{table*}[!ht]
\small
\begin{tcolorbox}[
  enhanced,
  colback=gray!5, colframe=gray!40,
  colbacktitle=gray!55, coltitle=white,
  fonttitle=\small\bfseries, fontupper=\small,
  title={LLM Judge Prompt Template},
  boxrule=0.8pt, arc=2pt,
  top=4pt, bottom=4pt, left=6pt, right=6pt,
  before upper={\setlength{\parskip}{2pt}\strut},
]

\textbf{\#\#\# Role}

You are a senior AI product evaluator specializing in assessing assistant
response quality under specific Skill scenarios.
Given the target Skill definition, evaluate the response against the ideal
behavior a fully competent reply should exhibit---not merely what the Skill
Body prescribes, but what would genuinely satisfy the user.

\vspace{0.5em}
\textbf{\#\#\# Input}

\textbf{Skill Body:}~\textcolor[rgb]{0,0,0.7}{\{Skill Body\}} \quad
\textbf{Response:}~\textcolor[rgb]{0,0,0.7}{\{Response\}}

\vspace{0.5em}
\textbf{\#\#\# Routing-Deviation Fallback Principle}
\quad\textit{(highest priority)}

When the input does not perfectly match the Skill's trigger scope
(\emph{routing deviation}), an \textbf{ideal response} must:
(1)~detect the deviation, (2)~infer the user's true intent, and
(3)~adaptively deliver a satisfying answer.
\textbf{Two behaviors are penalized equally}:
(a)~mechanically applying the Skill template while ignoring the user's real
need; and (b)~returning an error or silence without any useful fallback.
A reasonable rule deviation made to better serve user intent is \emph{not}
penalized---it should be credited.

\vspace{0.5em}
\textbf{\#\#\# Evaluation Dimensions}

\textbf{TCR --- Tool Call Compliance} [0--10]:
Are all necessary tools invoked in the correct order with accurate parameters?
\textit{Good} (8--10): complete, ordered, results used in answer;
\textit{Avg} (4--7): minor param gaps or underutilized results;
\textit{Poor} (0--3): missing first-turn tools or calls disconnected from output.

\textbf{CCC --- Card Orchestration Compliance} [0--10]:
Is the card type correct? Are cards placed inline at point-of-mention?
Are titles (${\leq}$14 chars) and search terms precise?
Are follow-up prompts context-relevant and diverse?
\textit{Good} (8--10): correct type, timely placement, precise title and search terms, valuable follow-ups;
\textit{Avg} (4--7): type correct but placement late, or generic titles/search terms, or templated follow-ups;
\textit{Poor} (0--3): wrong card type, missing required cards, banned title words, or no follow-up.

\textbf{CQ --- Content Quality} [0--20]:
\textit{Factual accuracy} [8]: conclusions correct, no hallucination;
\textit{Readability} [6]: mobile-friendly, concise, well-structured;
\textit{Intent understanding} [6]: hits the user's core need---in routing-deviation scenarios this sub-dimension carries the most weight; rigid rule-following that misses user need scores 0--2.
\textit{Good} (16--20): accurate, clean layout, deeply addresses user intent;
\textit{Avg} (8--15): mostly correct but lacks depth or has formatting issues;
\textit{Poor} (0--7): core error, cluttered layout, or intent completely missed.

\textbf{CA --- Constraint Adherence} [0--10]:
Are mandatory rules followed and prohibited actions avoided?
Reasonable deviations to better serve user intent in routing-deviation scenarios are \emph{not} treated as violations.
\textit{Good} (8--10): all mandatory rules and prohibitions respected; flexible fallback in routing-deviation cases;
\textit{Avg} (4--7): main rules followed but 1--2 gaps, or execution slightly mechanical;
\textit{Poor} (0--3): core prohibition violated, or rigid template application in routing-deviation scenario leaves user unserved.

\vspace{0.5em}
\textbf{\#\#\# Evidence Requirement}

Every entry in \texttt{rule\_violations} and \texttt{ideal\_gaps}
must cite a concrete excerpt from the response---no imagined details.

\vspace{0.5em}
\textbf{\#\#\# Output Format} \quad (strict JSON, no other text)

\begin{itemize}[leftmargin=1.2em,topsep=1pt,itemsep=0pt]
  \item \texttt{scores} --- TCR: [0--10], CCC: [0--10], CQ: [0--20], CA: [0--10]
  \item \texttt{tiers} --- Good / Average / Poor per dimension
  \item \texttt{total} --- sum of four scores (max 50)
  \item \texttt{rule\_violations} --- violations with evidence citations
  \item \texttt{ideal\_gaps} --- ideal behaviors absent from the response
  \item \texttt{skill\_md\_suggestions} --- Skill Body refinement patches
\end{itemize}

\end{tcolorbox}
\caption{LLM Judge prompt template used in Stage~3.
Raw scores (max 50) are linearly normalized to $[0,1]$ for reporting.}
\label{table:judge_prompt}
\end{table*}

\end{document}